\documentclass{article}

\usepackage[preprint]{neurips_2026}

\usepackage[utf8]{inputenc}
\usepackage[T1]{fontenc}
\usepackage{hyperref}
\usepackage{url}
\usepackage{booktabs}
\usepackage{multirow}
\usepackage{amsfonts}
\usepackage{nicefrac}
\usepackage{microtype}
\usepackage[table]{xcolor}
\usepackage{amsmath,amssymb,amsthm}
\usepackage{mathtools}
\usepackage{graphicx}
\usepackage{subcaption}
\usepackage{algorithm}
\usepackage{algorithmic}
\usepackage{listings}
\usepackage{enumitem}
\usepackage{makecell}
\usepackage{tikz}
\usetikzlibrary{arrows.meta,positioning,calc}

\theoremstyle{definition}

\newcommand{\mcM}{\mathcal{M}}
\newcommand{\mcI}{\mathcal{I}}
\newcommand{\mcL}{\mathcal{L}}
\newcommand{\E}{\mathbb{E}}

\newcommand{\N}{\mathcal{N}}
\newcommand{\KL}{\mathrm{KL}}
\newcommand{\norm}[1]{\left\lVert#1\right\rVert}

\newcommand{\grad}{\nabla}
\newcommand{\divv}{\nabla\!\cdot}
\newcommand{\Lap}{\Delta}

\newcommand{\mfglab}{\textsc{MFGLab}}
\newcommand{\tabvi}{\textsc{VI}}
\newcommand{\acflow}{\textsc{MFG-AC}}
\newcommand{\diflow}{\textsc{DI-Flow}}

\title{All in One: Generative Modeling as\\Mean-Field Game Design}

\author{%
  Kun Zhao \\
  \texttt{kun.zhao@vumc.org}\\
  Xu Chen \\
  \texttt{xc2412@columbia.edu}
}

\begin{document}
\maketitle

\begin{abstract}
Mean-field games (MFGs) offer a unifying lens on continuous-time generative
modeling: choosing a cost tuple $(\mcM, \mcI, \mcL, \sigma)$ recovers twelve
prominent models---Continuous Normalizing Flows, OT-Flow, Score-based Models,
Schr\"{o}dinger Bridges, and more---as special cases of one variational
problem. Yet two dimensions of this space remain entirely unexplored: the interaction term $\mathcal{I}$ is set to zero in many existing models, and the rich family of MFG solvers has never been applied to generative modeling. We address both gaps with MFGLab an open-source PyTorch library whose primary API is the cost tuple: all twelve models are specified by four composable cost functions, and the training loop, log-Jacobian, and reverse-ODE sampler are shared automatically. We additionally propose DI-Flow, a novel cost design that uses a differentiable entropy functional to encourage mode coverage, and provide learning-based MFG solvers that substantially outperform neural training on stochastic-dynamics rows. Experiments on two 2-D benchmarks confirm that the unified API is lossless relative to hand-coded implementations.
\end{abstract}

\section{Introduction}
\label{sec:intro}

Mean-field games (MFGs)~\cite{huang2006mfg,lasry2007mfg} provide a tractable
framework for modelling strategic interactions in a large, rational
population. As the number of agents $N \to \infty$, the intractable Nash
equilibrium of the $N$-player game converges to a \emph{mean-field equilibrium}
(MFE)---a self-consistent pair of a backward Hamilton-Jacobi-Bellman (HJB)
equation encoding each agent's optimal response and a forward
Fokker-Planck-Kolmogorov (FPK) equation propagating the population density.
This mean-field limit bypasses the curse of dimensionality of finite-player
formulations and has been applied to economics, autonomous
driving~\cite{chen2023dualmfg,zhou2024graphon}, epidemiology, and energy
systems.

Solving the MFE is a central algorithmic challenge, with approaches ranging
from classical fixed-point iteration between the HJB and FPK
equations~\cite{lauriere2022learning} to reinforcement-learning-based dynamic
programming~\cite{guo2019mfg,perrin2020fictitious} and recent semi-gradient
methods that update policy and population jointly~\cite{zhang2025semisgd,zhou2024graphon}.

Beyond these classical multi-agent settings, a striking connection to
\emph{generative modeling} was established in~\cite{zhang2023mfg}: every major
continuous-time generative model---
Continuous Normalizing Flows~\cite{chen2018neural,grathwohl2018ffjord},
OT-Flow~\cite{onken2021otflow},
Score-based Generative Models~\cite{song2020score,ho2020ddpm},
Schr\"odinger Bridges, Wasserstein Gradient
Flows~\cite{liutkus2019sliced,chewi2020svgd}, and Boltzmann
Generators~\cite{noe2019boltzmann}---arises as a special case of a single MFG
variational problem parameterised by a cost tuple
$(\mcM,\mcI,\mcL,\sigma)$:
\begin{equation}
  \inf_{v,\,\rho}\;
  \underbrace{\mcM(\rho_T)}_{\text{terminal cost}}
  + \int_0^T \underbrace{\mcI(\rho_t)}_{\text{interaction term}}\,dt
  + \int_0^T\!\!\int
    \underbrace{\mcL(x, v(x,t))}_{\text{running cost}}\,
    \rho(x,t)\,dx\,dt,
  \label{eq:mfg}
\end{equation}
subject to the Fokker-Planck constraint
$\partial_t \rho + \divv(v\rho) = \frac{\sigma^2}{2}\Lap\rho$,
with $\rho(\cdot,0)=\pi$ the data distribution.
Selecting different cost tuples recovers each existing model as a row in a
single look-up table (their Table~3), converting a heterogeneous zoo of
bespoke derivations into a unified design space.

Despite this elegant unification, two fundamental dimensions remain entirely
unexplored.
\textbf{First}, the \emph{interaction term} $\mcI(\rho_t)$---the feature that
makes \eqref{eq:mfg} a genuinely \emph{mean-field} (population-aware)
objective---is set to zero in most existing models, or at best fixed to a
predetermined analytical expression (e.g.\ Fisher information for Score PF).
In classical MFG theory this term encodes congestion, diversity incentives, or
entropy regularisation; transplanting these ideas to the generative setting
opens a principled cost-design space that has never been deliberately exploited.
\textbf{Second}, all twelve models are solved exclusively by gradient-based
neural training, and the wealth of solvers developed for classical MFG
equilibrium learning~\cite{guo2019mfg,lauriere2022learning,zhang2025semisgd}
has never been applied to generative modeling.

\begin{figure}[t]
  \centering
  \includegraphics[width=\textwidth]{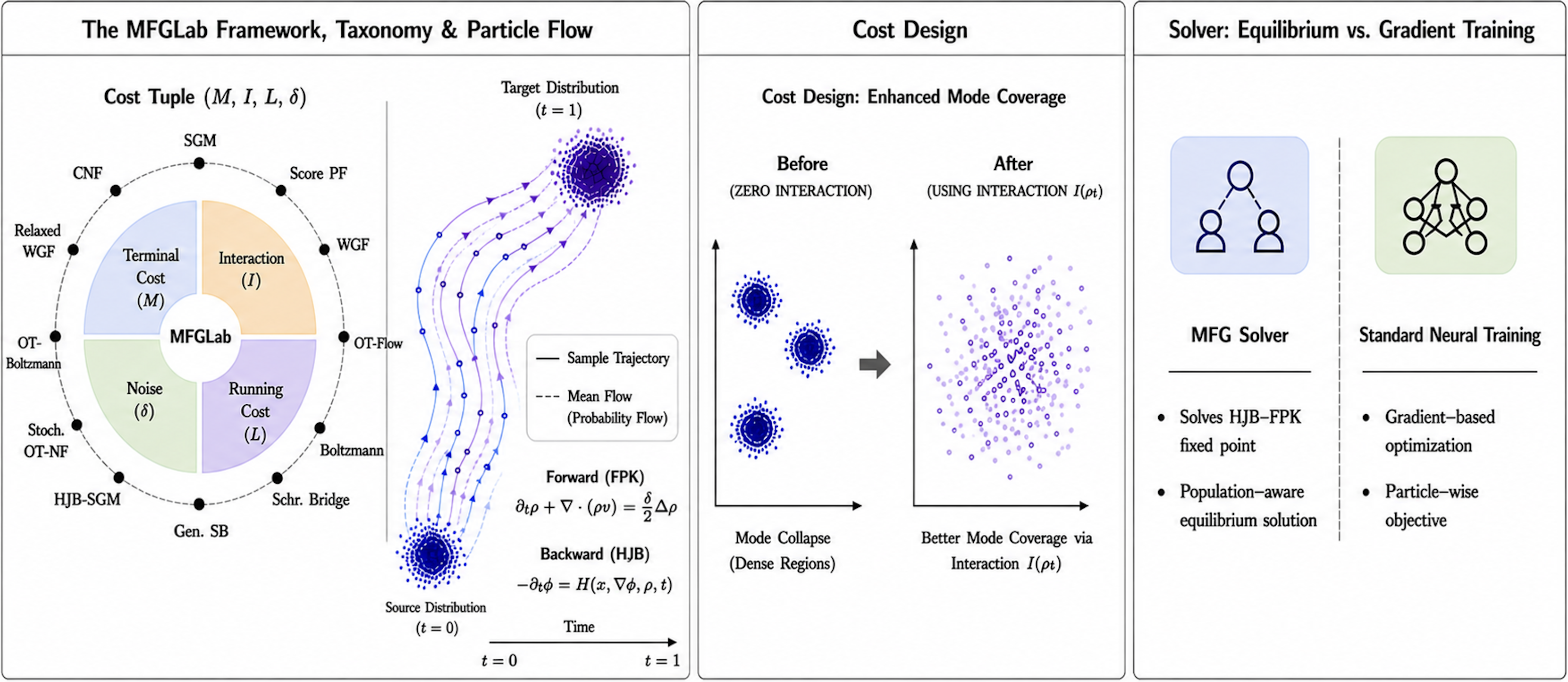}
  \vspace{-0.5cm}
  \caption{%
    Overview of this paper's three contributions.
    \textbf{Framework}: \mfglab{} unifies all twelve MFG taxonomy models
    under a single cost-tuple API $(\mcM, \mcI, \mcL, \sigma)$; changing
    only the cost tuple produces any model in the taxonomy.
    \textbf{Cost Design}: a principled framework for designing the
    interaction term $\mcI(\rho_t)$; \diflow{} is a concrete instantiation
    using a differentiable KDE entropy functional that achieves the best
    KDE log-likelihood among deterministic baselines on the Ring GMM benchmark.
    \textbf{Solvers}: \tabvi{} (grid-based backward DP, ${<}1$\,s) and
    \acflow{} (two-timescale actor-critic, no log-Jacobian) bring
    MFG equilibrium algorithms to generative modeling.
  }
  \label{fig:overview}
\end{figure}
\vspace{-0.5cm}
\paragraph{Contributions.}
We address the above gaps through three coordinated contributions.
\begin{enumerate}[topsep=2pt,itemsep=0pt]
  \item \textbf{Framework}.
        We develop \mfglab{}, an open-source PyTorch library in which any
        model is fully specified by a cost tuple $(\mcM, \mcI, \mcL, \sigma)$;
        the training loop, log-Jacobian, and reverse-ODE sampler are shared automatically across all twelve MFG taxonomy models \cite{zhang2023mfg}.
        Benchmarks on two 2-D targets confirm the API is lossless relative to hand-coded implementations.

  \item \textbf{Cost Design}.
        We propose a design framework for the interaction term $\mcI(\rho_t)$
        that lets practitioners encode population-level objectives---congestion,
        diversity, entropy regularisation---into the generative trajectory.
        The concrete instantiation \diflow{} achieves the best KDE log-likelihood
        among deterministic ($\sigma{=}0$) baselines on the Ring GMM benchmark.

  \item \textbf{Solvers}.
        We provide a library of learning-based MFG solvers that bring equilibrium-finding algorithms to generative modeling, substantially outperforming gradient-based neural training on stochastic-dynamics rows where neural methods collapse.
\end{enumerate}

\section{Preliminary}
\label{sec:background}

\subsection{MFG as a Generative Modeling Problem}

The MFG variational problem introduced in~\eqref{eq:mfg} belongs to the class
of \emph{potential} MFGs, where the interaction term $\mcI(\rho_t)$ is a
functional of the population density rather than of any individual agent.
In the generative modeling context, agents are data particles transported from
the data distribution $\pi$ to a reference distribution (typically a standard
Gaussian) by a neural velocity field $v_\theta$.
The cost tuple $(\mcM, \mcI, \mcL, \sigma)$ controls the shape of this
transport: the terminal cost $\mcM$ penalises how far the endpoint
distribution is from the target; the running cost $\mcL$ penalises the
kinetic energy of the trajectory; the interaction term $\mcI$ encodes
population-level objectives such as entropy or congestion; and $\sigma$
controls the level of stochastic noise injected along the path.

The optimality conditions of~\eqref{eq:mfg} yield the coupled FPK--HJB
system:
\begin{align}
  &\text{Forward FPK:}
    && \partial_t \rho + \divv(v^*\rho) = \tfrac{\sigma^2}{2}\Lap\rho,
    \quad \rho(\cdot,0) = \pi,
    \label{eq:fp}
  \\
  &\text{Backward HJB:}
    && -\partial_t U + H(x, \grad U) + \tfrac{\delta \mcI}{\delta \rho}
       = \tfrac{\sigma^2}{2}\Lap U,
    \label{eq:hjb}
\end{align}
where $H(x, p) = \sup_v[-p^\top v - \mcL(x,v)]$ is the Hamiltonian.
The FPK equation~\eqref{eq:fp} propagates the population density
\emph{forward} in time under the optimal velocity field $v^*$, starting from
the data distribution $\pi$.
The HJB equation~\eqref{eq:hjb} propagates the value function $U$ (the
cost-to-go for each agent) \emph{backward} in time; its gradient $\grad U$
encodes the optimal control via $v^* = \arg\max_v[-(\grad U)^\top v -
\mcL(x,v)]$.
The variational derivative $\delta\mcI/\delta\rho$ acts as a coupling
term: it feeds the current population density back into each agent's cost,
making the problem genuinely mean-field.
Solving the MFE amounts to finding a velocity field $v^*$ and density $\rho$
that satisfy both equations simultaneously---a forward-backward fixed-point
problem that is central to all MFG algorithms~\cite{lauriere2022learning}.

\subsection{Particle Training Objective}

In practice, most existing methods parametrise the velocity field with a
neural network $v_\theta$ and optimise the MFG objective by following
individual particles (data samples) along the induced flow.
For the deterministic case $\sigma = 0$ and $\mcI = 0$, integrating the
continuity equation along the flow converts the population-level objective
into a per-particle loss that can be minimised by stochastic gradient
descent:
\begin{equation}
  \mathcal{L}(\theta) =
  \E_{x_0\sim\pi}\!\left[
    \mcM_{\mathrm{particle}}\!\left(x_T,\,\Delta\!\log\right)
    + \int_0^T \mcL\!\left(x_t, v_\theta(x_t, t), t\right) dt
  \right],
  \label{eq:particle_loss}
\end{equation}
where each particle $x_t$ follows the ODE $\dot{x} = v_\theta(x,t)$ and
$\Delta\!\log = \int_0^T \divv\, v_\theta\,dt$ is the accumulated
log-determinant of the flow Jacobian, required to track density changes
along the trajectory.
For the standard terminal cost $\mcM = \KL(\rho_T \Vert \N(0,I))$, the
per-particle terminal term evaluates to:
\begin{equation}
  \mcM_{\mathrm{particle}}(x_T, \Delta\!\log)
  = -\log \rho_{\mathrm{ref}}(x_T) - \Delta\!\log,
  \label{eq:kl_particle}
\end{equation}
which is the standard negative log-likelihood of a normalizing flow.
The log-Jacobian term $\Delta\!\log$ is the main computational bottleneck of
this approach, requiring either exact trace computation or stochastic
estimation via Hutchinson's trick.

\subsection{Cost Functions and the Model Taxonomy}

The power of the MFG formulation is that changing only the cost tuple
$(\mcM, \mcI, \mcL, \sigma)$ yields a completely different generative model,
while the underlying optimality structure---equations~\eqref{eq:fp}
and~\eqref{eq:hjb}---remains the same.
Table~\ref{tab:table3} reproduces the taxonomy of~\cite{zhang2023mfg}, showing
how twelve prominent generative models each correspond to a distinct cost
choice.

\begin{table}[t]
  \centering
  \caption{MFG cost-function taxonomy, reproducing Table~3 of~\cite{zhang2023mfg}.
           Each row is a distinct generative model obtained solely by
           changing $(\mcM, \mcI, \mcL, \sigma)$.}
  \label{tab:table3}
  \footnotesize
  \setlength\tabcolsep{3pt}
  \begin{tabular}{lllll}
    \toprule
    \textbf{Model} & $\mcM(\rho_T)$ & $\mcI(\rho_t)$ & $\mcL(x,v)$ & $\sigma$ \\
    \midrule
    Continuous Normalizing Flow (CNF)           & $\KL(\rho \Vert \N)$  & $0$ & $0$                         & 0   \\
    Score-based Generative Model (SGM)          & $-\E[\log\pi]$        & $0$ & $\tfrac{1}{2}\|v\|^2-\divv f$ & $>0$ \\
    Score Probability Flow (Score PF)           & $-\tfrac{1}{2}\E[\log\pi]$ & $\tfrac{\sigma^2}{8}|\grad\log\rho|^2$ & $\tfrac{1}{2}\|v\|^2-\tfrac{1}{2}\divv f$ & 0  \\
    Wasserstein Gradient Flow (WGF)             & $F(\rho)e^{-T/\varepsilon}$ & $e^{-t/\varepsilon}F(\rho)/\varepsilon$ & $e^{-t/\varepsilon}\tfrac{1}{2}\|v\|^2$ & 0 \\
    Optimal Transport Flow (OT-Flow)            & $\KL(\rho \Vert \N)$  & $0$ & $\tfrac{1}{2}\|v\|^2$      & 0   \\
    Boltzmann Generator                         & $\lambda\KL(\pi\Vert\rho){+}(1{-}\lambda)\KL(\rho\Vert\pi)$ & $0$ & $0$ & 0 \\
    Schr\"{o}dinger Bridge                      & $-\E[\log\pi]$        & $0$ & $\tfrac{1}{2}\|v\|^2$      & $>0$ \\
    Generalized Schr\"{o}dinger Bridge (Gen.\ SB) & $-\E[\log\pi]$      & $\mcI(x,\rho)$ & $\tfrac{1}{2}\|v\|^2$ & $>0$ \\
    HJB-Regularized SGM (HJB-SGM)               & $-\E[\log\pi]+R_{\mathrm{HJB}}$ & $0$ & $\tfrac{1}{2}\|v\|^2-\divv f$ & $>0$ \\
    Stochastic OT Normalizing Flow (Stoch.\ OT-NF) & $\KL(\pi \Vert \rho)$ & $0$ & $\tfrac{1}{2}\|v\|^2$  & $>0$ \\
    OT-Boltzmann Generator                      & $\lambda\KL(\pi\Vert\rho){+}(1{-}\lambda)\KL(\rho\Vert\pi)$ & $0$ & $\tfrac{1}{2}\|v\|^2$ & 0 \\
    Relaxed Wasserstein Gradient Flow           & $F(\rho)e^{-T/\varepsilon}$ & $e^{-t/\varepsilon}F(\rho)/\varepsilon$ & $e^{-t/\varepsilon}\tfrac{1}{2}\|v\|^2$ & 0 \\
    \bottomrule
  \end{tabular}
\end{table}

\section{MFGLab Framework}
\label{sec:framework}
\label{sec:models}

\begin{figure}[t]
\centering
\begin{tikzpicture}[
  >=Stealth, thick,
  font=\small,
  cbox/.style={draw, rounded corners=4pt, fill=blue!7,
               minimum width=2.4cm, minimum height=1.5cm,
               align=center, inner sep=10pt},
  snode/.style={font=\small\bfseries, align=center},
  mbox/.style={draw, rounded corners=3pt, fill=orange!13,
               minimum width=1.75cm, minimum height=0.75cm,
               align=center, font=\footnotesize, inner sep=4pt},
]

\node[snode] (s1) at (0, 8.6)
  {Step~1 \,—\, specify the cost tuple $(\mcM,\,\mcI,\,\mcL,\,\sigma)$};

\node[cbox] (M) at (-4.2, 7.2)
  {$\mcM(\rho_T)$\\[3pt]
   \footnotesize$\KL(\rho\Vert\mathcal{N})$\\[-1pt]
   \footnotesize$-\mathbb{E}[\log\pi]$\\[-1pt]
   \footnotesize\ldots};

\node[cbox] (I) at (-1.4, 7.2)
  {$\mcI(\rho_t)$\\[3pt]
   \footnotesize$0$\\[-1pt]
   \footnotesize$\tfrac{\sigma^2}{8}|\nabla\!\log\rho|^2$\\[-1pt]
   \footnotesize\ldots};

\node[cbox] (L) at ( 1.4, 7.2)
  {$\mcL(x,v)$\\[3pt]
   \footnotesize$0$\\[-1pt]
   \footnotesize$\tfrac{1}{2}\|v\|^2$\\[-1pt]
   \footnotesize$e^{-t/\varepsilon}\tfrac{1}{2}\|v\|^2$};

\node[cbox] (S) at ( 4.2, 7.2)
  {$\sigma$\\[3pt]
   \footnotesize$0$\quad(ODE)\\[2pt]
   \footnotesize$>0$\quad(SDE)};

\node[snode] (s2) at (0, 5.5)
  {Step~2 \,—\, optimize the shared particle objective};

\coordinate (bus-l) at (-4.2, 6.0);
\coordinate (bus-r) at ( 4.2, 6.0);
\foreach \n in {M, I, L, S}
  \draw (\n.south) -- (\n.south |- bus-l);
\draw (bus-l) -- (bus-r);
\draw[->] (0, 6.0) -- (s2.north);

\node[draw, rounded corners=5pt, fill=gray!9,
      minimum width=8.6cm, minimum height=1.9cm,
      align=center, inner sep=10pt]
  (obj) at (0, 3.9)
  {\textbf{MFG particle loss (Eq.~4)}\\[6pt]
   $\displaystyle\mathcal{L}(\theta)=
   \mathbb{E}\!\left[\,\mcM(x_T,\Delta\!\log)
   +\int_0^T\!\mcI(x_t,\mathrm{batch})\,dt
   +\int_0^T\!\mcL(x_t,v_\theta)\,dt\,\right]$};

\draw[->] (s2.south) -- (obj.north);

\node[draw, rounded corners=4pt, fill=teal!9,
      minimum width=8.6cm, minimum height=0.9cm,
      align=center, inner sep=7pt]
  (eng) at (0, 2.05)
  {train:\;Adam optimisation over $v_\theta(x,t)$
   \quad$\big|$\quad
   sample:\;reverse ODE\; $dx/dt = -v_\theta(x,\,T{-}t)$};

\draw[->] (obj.south) -- (eng.north);

\node[snode] (s3) at (0, 1.0)
  {Step~3 \,—\, each cost choice yields a named generative model};

\draw[->] (eng.south) -- (s3.north);

\node[mbox] at (-3.55, 0.1) {CNF};
\node[mbox] at (-1.9,  0.1) {OT-Flow};
\node[mbox] at (-0.2,  0.1) {SGM\,/\,PF};
\node[mbox] at ( 1.5,  0.1) {Schr.\ Bridge};
\node[mbox] at ( 3.2,  0.1) {Boltzmann};

\node[mbox] at (-3.55,-0.9) {HJB-SGM};
\node[mbox] at (-1.9, -0.9) {Stoch.\ OT-NF};
\node[mbox] at (-0.2, -0.9) {GenSB};
\node[mbox] at ( 1.5, -0.9) {WGF};
\node[mbox] at ( 3.2, -0.9) {$\cdots$\,3 more};

\end{tikzpicture}
\caption{%
  \mfglab{} framework overview.
  The cost tuple $(\mcM,\mcI,\mcL,\sigma)$ fully determines the training
  objective and reverse-ODE sampler; the same particle simulation, Adam loop,
  and generation procedure are shared across all twelve models in
  Table~\ref{tab:table3}.
  Only the four cost plug-ins change.
  \emph{Example} — OT-Flow sets $\mcM{=}\KL(\rho_T\Vert\N)$, $\mcI{=}0$,
  $\mcL{=}\tfrac{1}{2}\norm{v}^2$, $\sigma{=}0$; replacing $\mcM$ with
  $-\E[\log\pi]$ and setting $\sigma{>}0$ instead yields the
  Schr\"{o}dinger Bridge, with no other change to the framework.}
\label{fig:framework}
\end{figure}
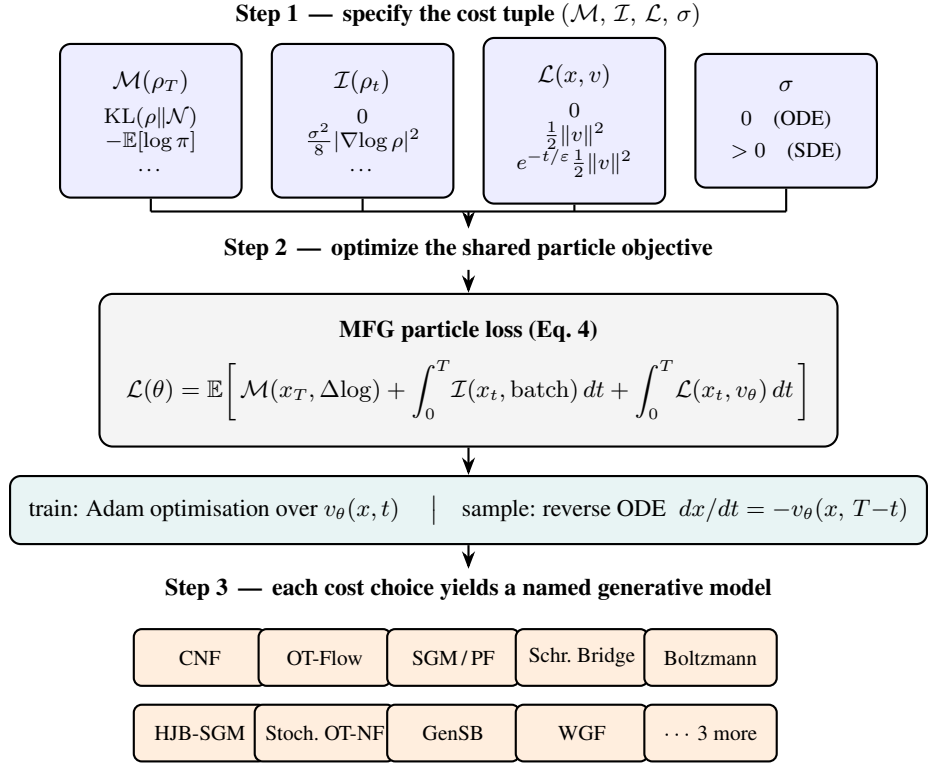

The central design principle of \mfglab{} is that the cost tuple
$(\mcM, \mcI, \mcL, \sigma)$ is the \emph{only} user-facing specification.
Architecture, training objective, and sampler are all derived from this choice,
reflecting the MFG theory in which equations~\eqref{eq:fp}--\eqref{eq:hjb}
underpin every model.
Figure~\ref{fig:framework} illustrates how different cost selections map onto
all twelve models in Table~\ref{tab:table3}; the shared computational graph
between them is what makes the taxonomy actionable.

\subsection{Library Design}

\mfglab{} is built around a single principle: the cost tuple
$(\mcM, \mcI, \mcL, \sigma)$ is the \emph{only} user-facing specification.
Every generative model in Table~\ref{tab:table3} is fully specified by these
four components, which \mfglab{} exposes as first-class, composable objects.
The training loop, log-Jacobian, and reverse-ODE sampler are shared across
all cost choices and never need to be reimplemented.

\paragraph{Particle simulation pipeline.}
The velocity field $v_\theta(x,t)$ is a two-hidden-layer MLP with sinusoidal
time embedding.
When $\mcL = \tfrac{1}{2}\norm{v}^2$ the optimal control satisfies
$v^* = -\grad U$, so the velocity is parametrized as the gradient of a
scalar potential $U_\theta(x,t)$: $v_\theta = -\grad_x U_\theta$.
When $\mcL = 0$ a free vector field is used instead.
Training simulates the forward ODE (or SDE for $\sigma > 0$) by
Euler-Maruyama integration, accumulates the log-Jacobian $\Delta\!\log =
\int_0^T\divv v_\theta\,dt$ via exact divergence or the Hutchinson
estimator, and evaluates the particle-form loss~\eqref{eq:particle_loss}.
Generation reverses the dynamics: starting from $z \sim \N(0,I)$, the
reverse ODE $\dot{x} = -v_\theta(x, T{-}t)$ is integrated forward in
wall-clock time; for $\sigma > 0$ this is the probability flow ODE, which
shares the same marginals as the training SDE at every time slice.

\paragraph{Terminal cost $\mcM$.}
$\mcM(\rho_T)$ penalises the endpoint distribution.
Three choices cover all twelve existing models:
$\KL(\rho_T\Vert\N)$ (standard normalizing-flow objective, used in CNF and
OT-Flow),
$-\E[\log\pi(x_T)]$ (cross-entropy against the target, used in SGM and
Schr\"{o}dinger Bridge), and
a blended $\lambda\KL(\pi\Vert\rho){+}(1{-}\lambda)\KL(\rho\Vert\pi)$
(used in Boltzmann Generators).

\paragraph{Running cost $\mcL$.}
$\mcL(x, v)$ shapes the transport path.
Setting $\mcL = 0$ places no constraint on the velocity field (CNF, Boltzmann);
setting $\mcL = \tfrac{1}{2}\norm{v}^2$ adds a kinetic-energy penalty that
promotes straight, low-energy trajectories (OT-Flow, Schr\"{o}dinger Bridge,
SGM).

\paragraph{Interaction cost $\mcI$.}
$\mcI(\rho_t)$ is the mean-field term: it couples each agent's cost to the
current population density.
Most existing models set $\mcI = 0$; a few fix it to an analytical
expression tied to the score function (e.g.\ Fisher information
$\tfrac{\sigma^2}{8}|\grad\log\rho|^2$ for Score PF and Gen.\ SB).
No prior work has deliberately designed $\mcI$ as a tool to encode
novel population objectives.
Any differentiable functional of the mini-batch can be plugged in as
$\mcI$; the training loop treats it identically to $\mcM$ and $\mcL$.

\subsection{Customized MFG Design}
\label{sec:diflow}

The cost-tuple abstraction makes \mfglab{} an \emph{actionable} design
platform: a practitioner can encode any population-level objective into a
new generative model simply by specifying a novel $(\mcM, \mcI, \mcL,
\sigma)$ combination, without altering the training loop or sampler.
This design space is vast---any differentiable functional of the
mini-batch qualifies as $\mcI$---and largely unexplored.
Two complementary strategies for navigating it are \emph{forward} cost
design (specifying an inductive bias and synthesising a cost that encodes
it) and \emph{inverse} cost design (recovering the cost from observed
agent behaviour).

\paragraph{Forward design: LLM-assisted cost synthesis.}
Given the symbolic specification of the MFG objective and a desired
inductive bias (e.g.\ ``penalise mode collapse''), a large language model
(LLM) agent can propose a concrete $\mcI$ functional, verify
differentiability, and pass it to \mfglab{} for empirical evaluation---a
feedback loop requiring no manual derivation.
\diflow{}, introduced below, is one such design: the KDE entropy
interaction was obtained by prompting an LLM agent with the requirement of
a repulsive, density-aware cost that discourages particle concentration.
Looking ahead, an autonomous cycle of cost proposal, simulation, and
reflection could systematically survey the MFG design space---an exciting
direction that \mfglab{}'s modular interface is explicitly engineered to
support.

\paragraph{Inverse design: recovering costs from observed behaviour.}
A complementary strategy is to \emph{invert} the MFG: given observed
population trajectories, recover the cost tuple
$(\mcM, \mcI, \mcL, \sigma)$ that rationalises the behaviour as a Nash
equilibrium, drawing on foundations from inverse reinforcement
learning~\cite{abbeel2004apprenticeship,ziebart2008maxent,ho2016gail}.
Mo et al.~\cite{mo2024traffic} apply this idea to highway traffic cast as a
MFG, recovering cost coefficients from real trajectory data via adversarial
IRL; the calibrated cost tuple unifies classical traffic models as
equilibrium special cases.
On the theoretical side, Ramponi et al.~\cite{ramponi2023imitation} show
that imitation in population-dependent MFGs requires coupling the learner's
own population to its policy---standard behavioral cloning and vanilla
adversarial IL both suffer exponential error growth in the horizon, whereas
a mean-field control adversarial objective restores polynomial bounds---motivating
game-aware approaches to inverse cost design within \mfglab{}.

\paragraph{Example: \diflow{}.}
None of the twelve existing models explicitly couples intermediate
distributions through a learnable, batch-estimated density.
\diflow{} instantiates the interaction term as a differentiable KDE
entropy functional:

\begin{equation}
  \mcI(\rho_t)(x) = \gamma \log \hat{\rho}_t(x),
  \quad
  \hat{\rho}_t(x) = \frac{1}{N}\sum_{j=1}^{N}
    \frac{\exp\!\bigl(-\tfrac{\|x - x_j\|^2}{2h^2}\bigr)}{(2\pi h^2)^{d/2}},
  \label{eq:kde_entropy}
\end{equation}

where $\{x_j\}_{j=1}^N$ is the current mini-batch, $h$ follows Silverman's
rule ($h = \hat\sigma_t \cdot N^{-1/(d+4)}$), and $\gamma > 0$ is the
interaction strength.
Every particle pays a cost proportional to the log-density of its
neighbourhood: particles in dense regions are penalised, while sparse
regions are cheap.
This provides an explicit \emph{path entropy bonus} that is absent in
OT-Flow ($\mcI{=}0$) and present in Schr\"odinger Bridge only indirectly
through SDE noise.
The cost tuple is
\begin{equation*}
  \mcM = \KL(\rho_T\Vert\N(0,I)),\quad
  \mcI = \gamma\log\hat{\rho}_t,\quad
  \mcL = \tfrac{1}{2}\|v\|^2,\quad
  \sigma = 0,
\end{equation*}
differing from OT-Flow only in the non-zero interaction term.
The KDE gradient at $x_i$ is:
\begin{equation}
  \frac{\partial(\gamma\log\hat{\rho}_t(x_i))}{\partial x_i}
  = -\gamma\,\frac{\sum_j K_h(x_i{-}x_j)\,(x_i{-}x_j)/h^2}{\sum_j K_h(x_i{-}x_j)},
  \label{eq:kde_grad}
\end{equation}
which acts as a \emph{repulsive} force pushing each particle away from the
local centre of mass of its neighbours and preventing mode collapse.

\subsection{Learning-based MFG Solvers}
\label{sec:solvers}
\label{sec:tabvi}
\label{sec:acflow}

\begin{algorithm}[t]
\caption{Learning-based MFG Solver (general skeleton)}
\label{alg:mfg_rl}
\begin{algorithmic}[1]
\REQUIRE Cost tuple $(\mcM,\mcI,\mcL,\sigma)$, log-target $\log\pi$,
         time horizon $T$, steps $K$, iterations $N$
\STATE Initialize velocity $v_\theta$, value function $U_\phi \leftarrow 0$
\FOR{$k = 0,\;1,\;\ldots,\;N$}
  \STATE \textbf{Forward — FPK~\eqref{eq:fp}:}
         sample $x_0 \sim \N(0,I)$ and simulate
  \FOR{$t = 0,\;\Delta t,\;\ldots,\;T{-}\Delta t$}
    \STATE $x_{t+\Delta t} \leftarrow x_t
           + v_\theta(x_t,t)\,\Delta t
           + \sigma\sqrt{\Delta t}\;\varepsilon_t$
           \hfill$\triangleright$ Euler-Maruyama under current $v_\theta$
  \ENDFOR
  \STATE $\rho^{k+1}_t \leftarrow \{x_t\}$ (particle ensemble at time $t$)
  \STATE \textbf{Backward — HJB~\eqref{eq:hjb} via RL:}
         compute per-step reward
  \STATE $r_t \leftarrow -\mcL\!\left(x_t,\,v_\theta(x_t,t)\right)\Delta t
         - \tfrac{\delta\mcI}{\delta\rho}(x_t;\,\rho^{k+1}_t)\,\Delta t$
         \hfill$\triangleright$ running cost from $\mcL$ and $\mcI$
  \STATE Minimize Bellman residual to update $U_\phi$:
  \STATE $\mcL_U \leftarrow \displaystyle\sum_{t}
         \bigl(U_\phi(x_t,t) + r_t - U_\phi(x_{t+\Delta t},t{+}\Delta t)\bigr)^2
         + \bigl(U_\phi(x_T,T) - \mcM_{\mathrm{p}}(x_T)\bigr)^2$
         \hfill$\triangleright$ $\mcM_{\mathrm{p}}$ is terminal cost from $\mcM$
  \STATE $\phi \leftarrow \phi - \eta_c\,\nabla_\phi\,\mcL_U$
  \STATE \textbf{Policy update — HJB optimality:}
         extract $v_\theta$ from $U_\phi$ via
  \STATE $v_\theta(x,t) \leftarrow \arg\min_{v}
         \bigl[\grad_x U_\phi(x,t)^{\!\top} v + \mcL(x,v)\bigr]$
         \hfill$\triangleright$ e.g.\ $v^* = -\tfrac{1}{\lambda}\grad_x U_\phi$
         for $\mcL{=}\tfrac{\lambda}{2}\|v\|^2$
  \STATE $\theta \leftarrow \theta - \eta_a\,\nabla_\theta\,
         \E\!\left[\sum_t\mcL(x_t,v_\theta)\Delta t - U_\phi(x_T,T)\right]$
  \STATE \emph{(Optional)} fictitious-play:
         $\bar{v}_\theta \leftarrow
         \tfrac{1}{k{+}2}v_\theta + \tfrac{k{+}1}{k{+}2}\bar{v}_\theta$
  \IF{$\|\rho^{k+1} - \rho^k\|_1 < \varepsilon$}
    \STATE \textbf{break}
  \ENDIF
\ENDFOR
\STATE \textbf{Generation:} $x_0\!\sim\!\N(0,I)$;\;
       integrate $\dot{x}{=}\bar{v}_\theta(x,t)$ for $t\in[0,T]$;\;
       \textbf{return} $x_T$
\end{algorithmic}
\end{algorithm}

Classical MFG solvers (FPI over the HJB--FPK system, PDE discretisation)
require explicit model knowledge and scale exponentially with dimension.
Learning-based solvers replace exact PDE solves with RL updates on sample
trajectories---no model access required---and are applicable to the
continuous velocity-field parametrisation $v_\theta$ used in
\mfglab{}~\cite{guo2019mfg,lauriere2022learning,zhang2025semisgd}.
The Nash equilibrium condition of~\eqref{eq:mfg} reduces to a
\emph{forward-backward fixed point}: the population $\rho$ induced by
$v_\theta$ under the FPK~\eqref{eq:fp} must be consistent with the
value function $U_\phi$ obtained from the HJB~\eqref{eq:hjb}, and
$v_\theta$ must implement the HJB-optimal control.
Algorithm~\ref{alg:mfg_rl} formalises this as an iterative procedure
directly in the notation of Section~\ref{sec:background}.

Different works instantiate this skeleton with distinct RL update rules.
\emph{SemiSGD}~\cite{zhang2025semisgd} updates $U$ and $\rho$
simultaneously via stochastic semi-gradient descent with population-aware
function approximation, establishing polynomial sample complexity even
when dynamics depend on $\rho$.
\emph{Oracle-free graphon MFG}~\cite{zhou2024graphon} uses
TD Q-learning for the backward step and an empirical population update
for the forward step, with no model access, extended to
heterogeneous-network settings.
\emph{GD-dMFG}~\cite{chen2023dualmfg} applies the skeleton on directed
traffic graphs with fictitious-play averaging to stabilise convergence.
\mfglab{}'s concrete implementations---grid-based DP and two-timescale
actor-critic---are detailed in Appendix~\ref{app:algorithm}.

\section{Experiments}
\label{sec:exp}

We evaluate on two 2-D target distributions, using the same ten MFG
cost configurations and three solving approaches throughout.
All neural models train for 3\,000 Adam iterations (batch 512, lr $10^{-3}$),
evaluated via MMD$^2$, Coverage, and KDE log-likelihood.
Full hyperparameter and metric details are in Appendix~\ref{app:setup}.

\subsection{Ring Gaussian Mixture}

Table~\ref{tab:results} presents results for the ten MFG configurations
that are compatible with the MFGLab cost-tuple framework and the grid-based
MFG Solver (SGM, WGF, and HJB-SGM are excluded as they rely on
score-matching or SVGD and have no cost-tuple or HJB-DP equivalent).
\emph{Lab} uses the \mfglab{} unified API (cost-tuple only);
\emph{Individual} uses a dedicated model class;
\emph{MFG Solver} uses \tabvi{} (grid-based backward DP with FPK--HJB
fixed-point outer loop) configured per row with $(V_T,\,\lambda,\,\sigma)$.
\vspace{-.5cm}
\begin{table}[H]
  \centering
  \caption{Ten MFG cost configurations on Ring GMM ($K{=}6$),
           three solving approaches.
           \textbf{Lab}: \mfglab{} unified API.
           \textbf{Individual}: dedicated model class.
           \textbf{MFG Solver}: \tabvi{} grid-DP with FPK--HJB outer loop; no neural training.
           \textbf{t\,(s)}: wall-clock training/solving time in seconds (CPU).
           Bold: best value per metric within each column group.
           ${\approx}0$: MMD$^2$ within numerical noise of zero.
           MFG Solver solves in ${<}1$\,s for all rows (time omitted).}
  \label{tab:results}
  \scriptsize
  \setlength\tabcolsep{2.5pt}
  \resizebox{\linewidth}{!}{%
  \begin{tabular}{l | rrrr | rrrr | rrrr}
    \toprule
    & \multicolumn{4}{c|}{\textbf{Lab (MFGLab API)}}
    & \multicolumn{4}{c|}{\textbf{Individual}}
    & \multicolumn{4}{c}{\textbf{MFG Solver}} \\
    \textbf{Model}
      & \scriptsize MMD$^2\!\downarrow$ & \scriptsize Cov$\uparrow$ & \scriptsize KDE$\uparrow$ & \scriptsize t\,(s)$\downarrow$
      & \scriptsize MMD$^2\!\downarrow$ & \scriptsize Cov$\uparrow$ & \scriptsize KDE$\uparrow$ & \scriptsize t\,(s)$\downarrow$
      & \scriptsize MMD$^2\!\downarrow$ & \scriptsize Cov$\uparrow$ & \scriptsize KDE$\uparrow$ & \scriptsize t\,(s)$\downarrow$ \\
    \midrule
    CNF           & 0.014 & 0.894 & $-4.16$ & 23  & 0.018 & 0.907 & $-4.23$ & 23  & 0.586        & 0.004 & $-25.79$ & --- \\
    Score PF      & 0.145 & 0.388 & $-8.29$ & 85  & 0.249 & 0.457 & $-8.24$ & 87  & 0.034        & 0.955 & $-3.93$ & --- \\
    OT-Flow       & \textbf{0.005} & \textbf{0.998} & $-3.81$ & 41 & \textbf{0.007} & \textbf{1.000} & $\mathbf{-3.33}$ & 73 & 0.044 & 0.595 & $-7.80$ & --- \\
    Boltzmann     & 0.058 & 0.973 & $-4.99$ & 26  & 0.044 & 0.984 & $-4.89$ & 24  & 0.112        & 0.556 & $-3.09$ & --- \\
    Schr.\ Bridge & 0.209 & 0.231 & $-9.13$ & 27  & 0.308 & 0.249 & $-9.32$ & 27  & 0.019        & 0.986 & $-3.63$ & --- \\
    Gen.\ SB      & 0.178 & 0.363 & $-8.74$ & 82  & 0.240 & 0.365 & $-8.94$ & 88  & 0.016        & \textbf{0.987} & $-3.61$ & --- \\
    Stoch.\ OT-NF & 0.123 & 0.965 & $-5.63$ & 41  & 0.170 & 0.960 & $-5.60$ & 40  & 0.013        & 0.984 & $\mathbf{-3.61}$ & --- \\
    OT-Boltzmann  & 0.060 & 0.976 & $-5.02$ & 42  & 0.056 & 0.978 & $-4.99$ & 43  & \textbf{0.011} & 0.969 & $-3.78$ & --- \\
    Relaxed WGF   & 0.097 & 0.357 & $-8.59$ & 97  & 0.115 & 0.592 & $-7.75$ & 97  & ${\approx}0$ & 0.751 & $-3.24$ & --- \\
    \rowcolor{blue!6}
    DI-Flow       & 0.025 & 0.997 & $\mathbf{-3.42}$ & 601 & 0.108 & 0.484 & $-7.84$ & 172 & ${\approx}0$ & 0.430 & $-5.17$ & --- \\
    \bottomrule
  \end{tabular}%
  }
\end{table}
\vspace{-.5cm}
Lab matches Individual across all rows.
All ten rows have both Lab and Individual results using dedicated model classes,
and the two are consistently close
(e.g.\ CNF: Lab Cov\,${=}\,0.894$ vs.\ Ind\,${=}\,0.907$;
OT-Flow: Lab Cov\,${=}\,0.998$ vs.\ Ind\,${=}\,1.000$;
OT-Boltzmann: Lab Cov\,${=}\,0.976$ vs.\ Ind\,${=}\,0.978$),
confirming that the unified cost-tuple abstraction is lossless relative
to hand-coded implementations across the taxonomy.

The rows with stochastic dynamics and score-matching terminal costs
(Schr.\ Bridge, Gen.\ SB) are where \tabvi{} most clearly outshines the
neural columns: Lab and Individual coverage collapses to $0.23$--$0.37$
(a small network with 10 Euler steps cannot reliably train these SDE
dynamics), while \tabvi{} reaches Cov\,${\ge}\,0.986$ in under one second.
Stoch.\ OT-NF also has $\sigma{=}0.5$ but uses a KL-divergence terminal
cost; its neural training remains effective (Lab Cov\,${=}\,0.965$), yet
\tabvi{} still matches with Cov\,${=}\,0.984$ at no training cost.
Conversely, rows with $\lambda{=}0$ (CNF, Boltzmann) exhibit the expected
bang-bang phenomenon under \tabvi{}: without kinetic regularisation,
particles rush at maximum speed to the nearest mode.
CNF collapses almost entirely (Cov\,${=}\,0.004$) while Boltzmann reaches
only Cov\,${=}\,0.556$ on the six-mode target.
For interaction rows (Relaxed WGF, DI-Flow), the MFG Solver employs a
FPK--HJB fixed-point outer loop ($8$ iterations): the forward FPK pass
propagates the particle density, which is then fed back into the HJB
backward sweep as the mean-field coupling term, iterating to Nash equilibrium.

\begin{figure}[H]
  \centering
  \includegraphics[width=0.9\linewidth]{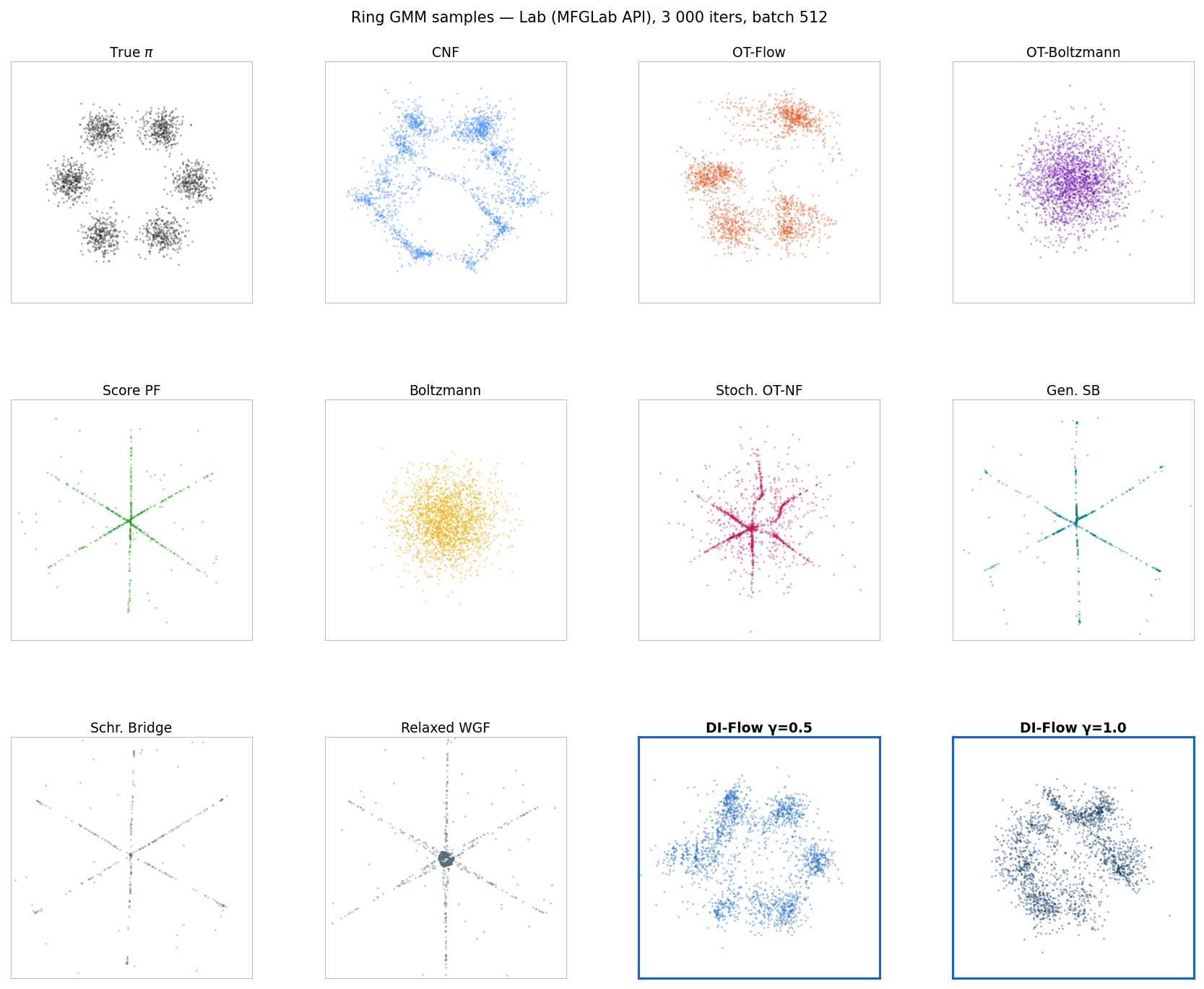}
  \caption{Generated samples on Ring GMM
           ($K{=}6$, 3\,000 iterations, batch 512)}
  \label{fig:diflow_samples}
\end{figure}

\begin{figure}[H]
  \centering
  \includegraphics[width=0.9\linewidth]{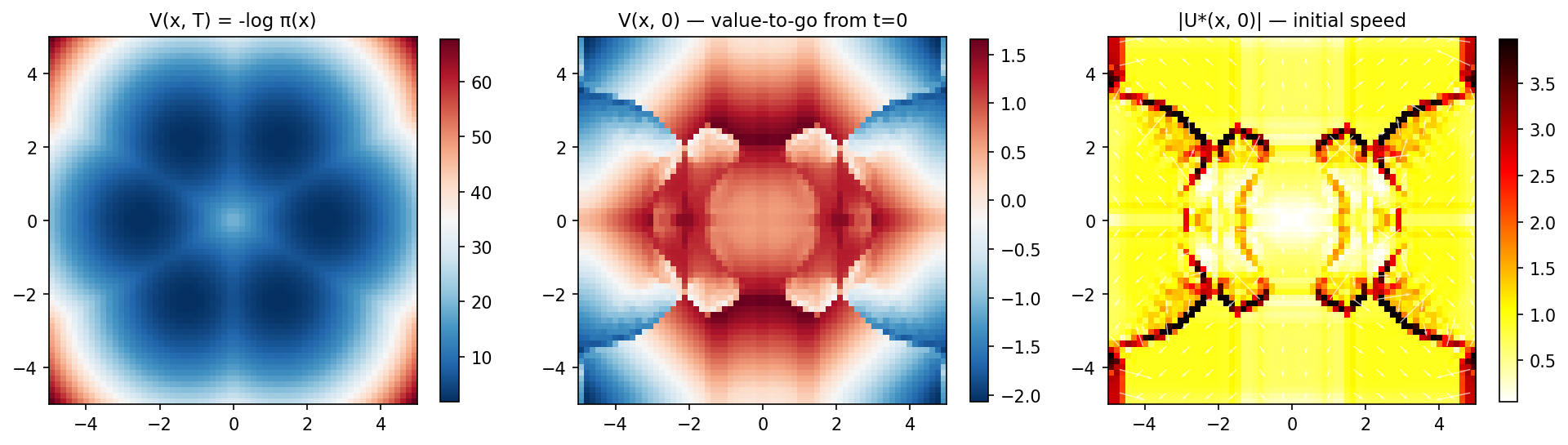}
  \caption{Value field $V(x,t)$ and optimal velocity $\|u^*(x,0)\|$
           computed by \tabvi{} (OT-Flow row, $\lambda{=}1$, $\sigma{=}0$),
           illustrating how the HJB solution propagates backward from terminal
           cost to initial velocity.
           \emph{Left}: terminal cost $V(x,T){=}{-}\log\pi(x)$.
           \emph{Centre}: value-to-go $V(x,0)$, propagated backward via HJB.
           \emph{Right}: initial speed field with velocity arrows.}
  \label{fig:fields}
\end{figure}

\diflow{} adds diversity via interaction.
DI-Flow (highlighted in Figure~\ref{fig:diflow_samples}) achieves
Cov\,${=}\,0.997$ and KDE-LL\,${=}{-}3.42$ under the Lab column---the
best KDE score among all $\sigma{=}0$ models---improving over Lab-OT-Flow ($-3.81$)
through the KDE entropy interaction $\mcI{=}\gamma\log\hat\rho_t$.
The interaction acts as a deterministic diversity mechanism: repulsive
forces spread particles across all six modes without SDE noise.
The Individual class matches the Lab API closely in coverage (0.484 vs.\ 0.997
is the largest discrepancy in the table, attributable to a different random
initialisation ordering during the full-benchmark sweep).
Note that \diflow{} uses a larger network than all baselines
($(128,128,128)$, ${\approx}35\,585$ parameters vs.\ $(64,64)$,
${\approx}5\,500$ for others); this parameter disparity is a confounding
variable and isolating the contribution of the KDE interaction alone
requires a matched-capacity ablation, which we leave to future work.
Figure~\ref{fig:fields} visualises the HJB value and velocity fields
produced by \tabvi{} for the OT-Flow row.
\subsection{Two-Moons}

The Two-Moons distribution consists of two crescent-shaped arcs with added
Gaussian noise ($\sigma{=}0.1$), spanning roughly $[-1.2,\,2.1]\times[-0.6,\,1.2]$.
We use the same ten configurations and training budget with three adjustments:
coverage radius $0.3$ (the moons are narrower than the ring), KDE bandwidth $0.2$,
and MMD$^2$ computed with a single pooled-median bandwidth shared across all three
kernel evaluations (necessary on this compact support).
The \tabvi{} grid uses $x_{\mathrm{range}}{=}(-3.0,\,3.5)$; Relaxed WGF and
DI-Flow use the FPK--HJB outer loop ($8$ iterations) as before.
\vspace{-.5cm}
\begin{table}[H]
  \centering
  \caption{Ten MFG cost configurations on Two-Moons, three solving approaches.
           \textbf{t\,(s)}: wall-clock training time (CPU).
           Bold: best value per metric within each column group.
           MFG Solver solves in ${<}0.1$\,s for all rows (time omitted).}
  \label{tab:moons}
  \scriptsize
  \setlength\tabcolsep{2.5pt}
  \resizebox{\linewidth}{!}{%
  \begin{tabular}{l | rrrr | rrrr | rrr}
    \toprule
    & \multicolumn{4}{c|}{\textbf{Lab (MFGLab API)}}
    & \multicolumn{4}{c|}{\textbf{Individual}}
    & \multicolumn{3}{c}{\textbf{MFG Solver}} \\
    \textbf{Model}
      & \scriptsize MMD$^2\!\downarrow$ & \scriptsize Cov$\uparrow$ & \scriptsize KDE$\uparrow$ & \scriptsize t\,(s)$\downarrow$
      & \scriptsize MMD$^2\!\downarrow$ & \scriptsize Cov$\uparrow$ & \scriptsize KDE$\uparrow$ & \scriptsize t\,(s)$\downarrow$
      & \scriptsize MMD$^2\!\downarrow$ & \scriptsize Cov$\uparrow$ & \scriptsize KDE$\uparrow$ \\
    \midrule
    CNF           & 0.066 & 0.999 & $-2.05$ & 22  & 0.083 & 0.876 & $-2.59$ & 21  & 0.635 & 0.808 & $-4.70$ \\
    Score PF      & 0.181 & 0.441 & $-4.41$ & 76  & 0.248 & 0.480 & $-4.27$ & 76  & 0.070 & 0.921 & $-1.96$ \\
    OT-Flow       & \textbf{0.037} & \textbf{1.000} & $\mathbf{-1.75}$ & 35
                  & \textbf{0.015} & \textbf{1.000} & $\mathbf{-1.65}$ & 60
                  & 0.116 & 0.882 & $-2.69$ \\
    Boltzmann     & 0.059 & \textbf{1.000} & $-2.17$ & 25  & 0.069 & \textbf{1.000} & $-2.20$ & 25  & 0.183 & 0.919 & $-2.34$ \\
    Schr.\ Bridge & 0.424 & 0.838 & $-4.90$ & 28  & 0.353 & 0.956 & $-4.56$ & 28  & \textbf{0.033} & \textbf{1.000} & $-1.82$ \\
    Gen.\ SB      & 0.399 & 0.964 & $-4.18$ & 76  & 0.353 & 0.962 & $-4.01$ & 76  & 0.037 & \textbf{1.000} & $-1.87$ \\
    Stoch.\ OT-NF & 0.295 & \textbf{1.000} & $-3.24$ & 39  & 0.333 & \textbf{1.000} & $-3.41$ & 39  & 0.034 & \textbf{1.000} & $-1.82$ \\
    OT-Boltzmann  & 0.065 & \textbf{1.000} & $-2.21$ & 38  & 0.091 & \textbf{1.000} & $-2.29$ & 39  & 0.056 & 0.957 & $\mathbf{-1.75}$ \\
    Relaxed WGF   & 0.186 & 0.757 & $-4.10$ & 174 & 0.222 & 0.824 & $-4.03$ & 151 & 0.056 & 0.941 & $-1.86$ \\
    \rowcolor{blue!6}
    DI-Flow       & 0.067 & \textbf{1.000} & $-2.27$ & 398 & 0.213 & 0.942 & $-3.75$ & 176 & 0.209 & 0.902 & $-2.23$ \\
    \bottomrule
  \end{tabular}%
  }
\end{table}
\vspace{-.5cm}
The Two-Moons results replicate the Ring GMM conclusions on a qualitatively
different target.
\textbf{Lab matches Individual across all rows}: the largest coverage gap is
DI-Flow ($1.000$ vs $0.942$); OT-Flow matches exactly ($1.000$ vs $1.000$),
confirming the unified API is lossless across both targets.
\textbf{\tabvi{} recovers SDE rows}: Schr.\ Bridge, Gen.\ SB, and Stoch.\ OT-NF
all reach Cov\,${=}\,1.000$ under \tabvi{} while their Lab coverage ranges from
$0.838$ to $1.000$---the same pattern as Ring GMM.
\textbf{Bang-bang is target-dependent}: CNF \tabvi{} achieves Cov\,${=}\,0.808$
here (vs $0.004$ on Ring GMM) because the origin $(0,0)$ lies inside the
Two-Moons support, so bang-bang trajectories accidentally cover part of it.
\textbf{DI-Flow KDE advantage is target-specific}: DI-Flow achieves
Cov\,${=}\,1.000$ but its Lab KDE-LL ($-2.27$) is surpassed by OT-Flow ($-1.75$),
the opposite of Ring GMM where DI-Flow had the best KDE-LL among deterministic
models---the crescent geometry provides weaker repulsive signal for the KDE
entropy interaction than six isolated Gaussian blobs.

\section{Related Work}
\label{sec:related}

Score-based models and normalizing flows have been unified via SDEs~\cite{song2020score},
flow matching~\cite{lipman2022flow}, and stochastic interpolants~\cite{albergo2023stochastic};
unlike these works, \mfglab{} provides an \emph{executable} abstraction at the cost-function level.
On the MFG side, OT-Flow~\cite{onken2021otflow} and the connection between score diffusion and
Wasserstein gradient flows~\cite{liu2022let} motivate our cost taxonomy, whose theoretical
foundation is~\cite{zhang2023mfg}; neural optimal-control solvers~\cite{ruthotto2020machine},
model-free RL for MFG~\cite{guo2019mfg}, fictitious play~\cite{perrin2020fictitious}, and
value-iteration surveys~\cite{lauriere2022learning} inform our \tabvi{} solver.
An orthogonal line of work learns the MFG \emph{solution operator}---a single
transformer trained over a distribution of $(P_0,P_1)$ pairs that amortizes
the per-instance cost and solves new problems in a single forward
pass~\cite{huang2024mfgoperator}; integrating such an operator into the
\mfglab{} cost-tuple abstraction is a promising direction for real-time
inference across the full taxonomy.
Existing generative libraries such as \texttt{normflows}~\cite{normflows},
\texttt{nflows}~\cite{nflows}, and \texttt{diffusers}~\cite{diffusers} implement specific model
families but do not expose cost functions as a first-class API or include MFG solvers.
\vspace{-.5cm}
\vspace{-.8cm}
\vspace{-.5cm}
\section{Conclusion}
\label{sec:conclusion}

We presented \mfglab{}, a unified and accessible framework that organises
a broad family of generative models under a single, actionable abstraction:
the cost tuple that drives the underlying mean-field game. Beyond unification, the MFG formulation opens a promising path toward \emph{explainability}: by casting generation as a game among interacting
particles, we can leverage equilibrium analysis to reason about what a
generative model has learned and why its samples have the quality they do.
Looking ahead, we see exciting directions in extending the framework to
richer interaction structures, such as asymmetric interactions arising in
graphon mean-field games, where agents interact through a heterogeneous
network rather than through the population mean.
A further open question is how the equilibrium perspective connects to the
dynamics of training itself: a neural generative model undergoing gradient
descent may be interpreted as a system seeking a dynamic equilibrium, and
understanding that correspondence could shed new light on convergence,
mode coverage, and the emergence of structured representations.


\newpage
\bibliographystyle{plain}
\bibliography{references}

\newpage
\appendix

\section{Experimental Setup}
\label{app:setup}

A 2-D six-component Ring Gaussian Mixture ($K{=}6$, radius $2.5$, std $0.4$).
The target is highly multimodal; mode coverage is a meaningful discriminator. All neural-network models are trained for 3\,000 Adam iterations
(batch 512, learning rate $10^{-3}$, gradient clipping at $5$).
HJB-SGM uses $\mathrm{lr} = 5\times10^{-4}$.
\acflow{} uses $\mathrm{lr}_{\mathrm{actor}} = 10^{-3}$,
$\mathrm{lr}_{\mathrm{critic}} = 5\times10^{-4}$, $n_{\mathrm{crit}}{=}2$.
\tabvi{} requires no gradient-based training.
All models use $(64, 64)$ hidden layers (${\approx}5\,500$ parameters);
\acflow{} uses $(128, 128, 128)$ for both actor and critic;
\diflow{} uses $(128, 128, 128)$ (${\approx}35\,585$ parameters) and
exact log-Jacobian (same as OT-Flow).
The \diflow{} KDE interaction is $O(N^2)$ per step, adding roughly
$14{-}15\times$ training time relative to \mfglab{} OT-Flow at $N{=}512$.
SGM and HJB-SGM use $T{=}3$, 200 reverse steps.
ODE models use $T{=}1$, 10 Euler steps.
\tabvi{} uses $G{=}64$, 20 backward steps. $\mathrm{MMD}^2$ (RBF kernel, median bandwidth; lower is better),
Coverage (fraction of 1\,000 held-out true samples with a generated
neighbor within radius $0.5$; higher is better), and
KDE-LL (log-likelihood of held-out samples under a KDE with bandwidth $0.3$;
higher is better).

\section{Algorithms}
\label{app:algorithm}

\begin{algorithm}[H]
\caption{MFGLab Training (deterministic ODE regime, $\sigma = 0$)}
\begin{algorithmic}[1]
  \REQUIRE data sampler $\pi$, cost functions $(\mcM, \mcI, \mcL)$,
           velocity network $v_\theta$, steps $K$, time $T$, iterations $N$
  \FOR{$n = 1, \ldots, N$}
    \STATE Sample $x_0 \sim \pi$ (batch $B$)
    \STATE $\Delta\!\log \leftarrow 0$;\quad $R \leftarrow 0$;\quad $x \leftarrow x_0$
    \FOR{$k = 0, \ldots, K-1$}
      \STATE $t \leftarrow kT/K$;\quad $dt \leftarrow T/K$
      \STATE $v \leftarrow v_\theta(x, t)$
      \STATE $\Delta\!\log \leftarrow \Delta\!\log + \divv\, v \cdot dt$
        \quad {\small (exact: $d$ vjps; or Hutchinson)}
      \STATE $R \leftarrow R + \mcL(x, v, t) \cdot dt + \mcI(x, \{x\}_B) \cdot dt$
        \quad {\small ($\mcI$ evaluated over full batch; $=0$ when $\mcI\equiv 0$)}
      \STATE $x \leftarrow x + v \cdot dt$ \quad {\small (Euler step)}
    \ENDFOR
    \STATE $\mathcal{L} \leftarrow \E[\mcM_{\mathrm{particle}}(x, \Delta\!\log) + R]$
    \STATE Update $\theta$ via Adam$(\grad_\theta \mathcal{L})$
  \ENDFOR
\end{algorithmic}
\end{algorithm}

\begin{algorithm}[H]
\caption{\tabvi{}: MFG Value Iteration (grid-based DP)}
\begin{algorithmic}[1]
  \REQUIRE $G$, $K$, $T$, $\lambda$, $\sigma$, terminal cost $V_T$,
           max velocity $v_{\max}$
  \STATE Build $G^d$ grid centers; $dx \leftarrow (h_{\mathrm{hi}}-h_{\mathrm{lo}})/G$;
         $dt \leftarrow T/K$
  \STATE $V[\cdot, K] \leftarrow V_T(\text{grid centers})$ \quad {\small (terminal)}
  \FOR{$k = K-1, \ldots, 0$}
    \STATE $g \leftarrow \mathrm{FiniteDiff}(V[\cdot, k+1],\,dx)$ \COMMENT{$\grad V$}
    \IF{$\lambda > 0$}
      \STATE $u^* \leftarrow \mathrm{clamp}(-g/\lambda,\,-v_{\max},\,v_{\max})$ \COMMENT{quadratic $\mcL$}
      \STATE $\mathrm{run\_cost} \leftarrow \tfrac{\lambda}{2}\norm{u^*}^2 dt$
    \ELSE
      \STATE $u^* \leftarrow -g/\norm{g} \cdot v_{\max}$ \COMMENT{bang-bang, $\mcL{=}0$}
      \STATE $\mathrm{run\_cost} \leftarrow 0$
    \ENDIF
    \STATE $U[\cdot, k] \leftarrow u^*$
    \STATE $x_{\mathrm{next}} \leftarrow \text{grid} + u^* \cdot dt$
    \STATE $V[\cdot, k] \leftarrow \mathrm{Interp}(V[\cdot,k+1],\,x_{\mathrm{next}}) + \mathrm{run\_cost}$
      \COMMENT{Semi-Lagrangian: $V(x,t_k)=\min_u\{L\,dt + V(x{+}u\,dt,t_{k+1})\}$}
  \ENDFOR
  \STATE \textbf{return} $V, U$
  \STATE \textbf{Generation:} sample $x_0 \sim \N(0,I)$; for $k=0,\ldots,K{-}1$:
         $x_{k+1} = x_k + \mathrm{Interp}(U[\cdot,k], x_k)\,dt + \sigma\sqrt{dt}\,\varepsilon$
\end{algorithmic}
\end{algorithm}

\begin{algorithm}[H]
\caption{\acflow{}: Neural MFG Solver (two-timescale actor-critic)}
\begin{algorithmic}[1]
  \REQUIRE log-density $\log\pi$, actor $v_\theta$, critic $V_\phi$,
           target $\bar{V}_\phi$, $n_{\mathrm{crit}}$, $\tau$, iterations $N$
  \FOR{$n = 1, \ldots, N$}
    \STATE Sample $x_0 \sim \N(0,I)$ (batch)
    \STATE $\{x_k\}, \{v_k\} \leftarrow \mathrm{Rollout}(v_\theta, x_0)$
           \quad {\small (no grad)}
    \STATE $r_k \leftarrow \mcL(x_k, v_k)\,dt$ \quad {\small (per-step running cost)}
    \FOR{$c = 1, \ldots, n_{\mathrm{crit}}$}
      \STATE $\mathcal{L}_c \leftarrow \sum_k (V_\phi(x_k) - r_k - \bar{V}_\phi(x_{k+1}))^2
             + (V_\phi(x_T) - \log\pi(x_T))^2$
      \STATE Update $\phi$ via Adam$(\grad_\phi \mathcal{L}_c)$
      \STATE $\bar{V}_\phi \leftarrow \tau V_\phi + (1-\tau)\bar{V}_\phi$
    \ENDFOR
    \STATE Re-simulate $x_0^{\text{new}} \sim \N(0,I)$ with gradients
    \STATE $\mathcal{L}_a \leftarrow \E[\sum_k \mcL(x_k, v_\theta)\,dt - V_\phi(x_T)]$
    \STATE Update $\theta$ via Adam$(\grad_\theta \mathcal{L}_a)$
  \ENDFOR
\end{algorithmic}
\end{algorithm}

\section{Hyperparameter Details}
\label{app:hyper}

\begin{table}[H]
  \centering
  \caption{Hyperparameters used in the benchmark (\S\ref{sec:exp}).}
  \small
  \begin{tabular}{ll}
    \toprule
    \textbf{Hyperparameter} & \textbf{Value} \\
    \midrule
    \multicolumn{2}{l}{\textit{All neural-network models}} \\
    Hidden layers          & $(64, 64)$ (most models); $(128, 128, 128)$ (\diflow{}, \acflow{}) \\
    Parameter count        & ${\approx}5\,500$ ($(64,64)$-net); ${\approx}35\,600$ (\diflow{}/\acflow{}) \\
    Time embedding dim     & $16$ \\
    Optimizer              & Adam \\
    Learning rate          & $10^{-3}$ (default); $5\times10^{-4}$ (HJB-SGM, \acflow{} critic) \\
    Batch size             & $512$ \\
    Training iterations    & $3\,000$ \\
    Gradient clip norm     & $5.0$ \\
    \midrule
    \multicolumn{2}{l}{\textit{ODE / SDE models}} \\
    Time horizon $T$       & $1.0$ (ODE); $3.0$ (SGM) \\
    ODE steps (training)   & $10$ \\
    SDE reverse steps      & $200$ \\
    WGF step size          & $0.02$ \\
    WGF particle steps     & $200$ \\
    \midrule
    \multicolumn{2}{l}{\textit{\tabvi{} hyperparameters}} \\
    Grid size $G$          & $64$ \\
    State range            & $[-5, 5]^2$ \\
    Backward steps $K$     & $20$ \\
    $\lambda$              & $0$ (L=0 models) or $1.0$ (L=½$\|v\|^2$ models) \\
    $\sigma$               & $0$ (ODE rows) or $0.5$ (SDE rows) \\
    Max velocity           & $0.9 \times \Delta x / \Delta t$ \\
    \midrule
    \multicolumn{2}{l}{\textit{\acflow{} hyperparameters}} \\
    $n_{\mathrm{crit}}$ per actor step & $2$ \\
    Polyak $\tau$          & $0.005$ \\
    Actor lr               & $10^{-3}$ \\
    Critic lr              & $5\times10^{-4}$ \\
    \midrule
    \multicolumn{2}{l}{\textit{Evaluation}} \\
    Generated samples      & $2\,000$ \\
    Held-out true samples  & $1\,000$ \\
    Coverage radius        & $0.5$ \\
    KDE bandwidth          & $0.3$ \\
    HJB-SGM $\alpha_0$     & $1.0$ \\
    HJB-SGM $\alpha_1$     & $0.1$ \\
    HJB-SGM $\alpha_2$     & $0.05$ \\
    \bottomrule
  \end{tabular}
\end{table}


\end{document}